\documentclass[5p,final,authoryear,twocolumn]{elsarticle}

\usepackage[utf8]{inputenc}
\usepackage{amssymb}
\usepackage{amsmath}
\usepackage{multirow}
\usepackage[hidelinks,breaklinks]{hyperref}
\usepackage[table]{xcolor}

\usepackage{bm}
\usepackage{xspace}
\usepackage{amssymb}
\usepackage{amsmath}
\usepackage{amsfonts}
\usepackage{euscript}

\newcommand{\bbk}{{\bf k}}

\newcommand{\bbr}{{\bf r}}

\newcommand{\bbw}{{\bf w}}

\newcommand{\bbX}{{\bf X}}
\newcommand{\bbZ}{{\bf Z}}

\newcommand{\bbW}{{\bf W}}

\newcommand{\mY}{{\mathcal Y}}
\newcommand{\mM}{{\mathcal M}}
\newcommand{\mN}{{\mathcal N}}

\newcommand{\bzeta}{\hbox{\boldmath $\zeta$}}

\newcommand{\bbeta}{\hbox{\boldmath $\beta$}}

\def\registered{{\ooalign {\hfil\raise .05ex\hbox{\scriptsize
R}\hfil\crcr\mathhexbox20D}}}

\def\REgistered{{\ooalign
{\hfil\raise.09ex\hbox{\tiny \sf R}\hfil\crcr\mathhexbox20D}}}

\makeatletter
\DeclareRobustCommand\onedot{\futurelet\@let@token\@onedot}
\def\@onedot{\ifx\@let@token.\else.\null\fi\xspace}

\graphicspath{{figs/}}

\usepackage{booktabs}
\newcommand{\revision}[1]{#1}

\begin{document}
\begin{frontmatter}

\title{Detect and Correct Bias in Multi-Site Neuroimaging Datasets} %: \\ Name That Dataset, Causal Inference, Harmonization}
%Quantifying Confounding Bias in Neuroimaging Datasets with Causal Inference}

\author{Christian Wachinger$^{1}$\corref{cor1}}  
\cortext[cor1]{Corresponding Author. Address: Waltherstr. 23, 80337 München, Germany; Email: christian.wachinger@med.uni-muenchen.de}
\author{Anna Rieckmann$^{2}$, Sebastian Pölsterl$^{1}$ \\ \scriptsize{for the Alzheimer's Disease Neuroimaging Initiative\corref{cor2} \\and the Australian Imaging Biomarkers and Lifestyle flagship study of ageing\corref{cor3}}}
\cortext[cor2]{Data used in preparation of this article were obtained from the Alzheimer's Disease Neuroimaging Initiative (ADNI) database (adni.loni.usc.edu). As such, the investigators within the ADNI contributed to the design and implementation of ADNI and/or provided data but did not participate in analysis or writing of this report.
A complete listing of ADNI investigators can be found at:
\url{http://adni.loni.usc.edu/wp-content/uploads/how_to_apply/ADNI_Acknowledgement_List.pdf}}
\cortext[cor3]{Data used in the preparation of this article was obtained from the Australian Imaging Biomarkers and Lifestyle flagship study of ageing (AIBL) funded by the Commonwealth Scientific and Industrial Research Organisation (CSIRO) which was made available at the ADNI database (www.loni.usc.edu/ADNI). The AIBL researchers contributed data but did not participate in analysis or writing of this report. AIBL researchers are listed at www.aibl.csiro.au.
}
\address{
$^1$Lab for Artificial Intelligence in Medical Imaging (AI-Med), Department of Child and Adolescent Psychiatry, University Hospital, LMU München, Germany \\
$^2$Ume{\aa} Center for Functional Brain Imaging, Department of Radiation Sciences, Ume{\aa} University, Sweden
}

\begin{abstract}
The desire to train complex machine learning algorithms and to increase the statistical power in association studies drives neuroimaging research
to use ever-larger datasets. 
The most obvious way to increase sample size is by pooling scans from  independent studies.
However, simple pooling is often ill-advised as selection, measurement, and confounding biases may creep in and yield spurious correlations. 
In this work, we combine 35,320 magnetic resonance images of the brain from 17 studies to examine bias in neuroimaging. 
In the first experiment, \emph{Name That Dataset}, we provide empirical evidence for the presence of bias by showing that  scans can be
correctly assigned to their respective dataset with 71.5\% accuracy.
Given such evidence, we take a closer look at confounding bias, which is often viewed as the main shortcoming in observational studies. 
In practice, we neither know all potential confounders nor do we have data on them. 
Hence, we model confounders as unknown, latent variables. 
Kolmogorov complexity is then used to decide whether the confounded or the causal model provides the simplest factorization of the graphical model. 
Finally, we present methods for dataset harmonization and study their ability to remove bias in imaging features. 
In particular, we propose an extension of the recently introduced ComBat algorithm to control for global variation across image features, inspired by adjusting for unknown population stratification in genetics. 
Our results demonstrate that harmonization can reduce dataset-specific information in image features. 
Further, confounding bias can be reduced and even turned into a causal relationship. 
However, harmonization also requires caution as it can easily remove relevant subject-specific information. 
Code is available at \url{https://github.com/ai-med/Dataset-Bias}.
%\todo[inline]{We switch between dataset/subject-specific \emph{with} and \emph{without} hyphen. We should pick one and stick with it.}
%First, we \emph{detect} bias by experimentally showing that scans can be
%correctly assigned to their respective dataset with 71.5\% accuracy.
%Second, we propose to tell causal from confounding factors by \emph{quantifying} the extent of confounding and causality in a single dataset using causal inference. 
%As we neither know all potential confounders nor do we have data on them, we model the confounder as unknown, latent variable.  
%Kolmogorov complexity is then used to decide whether the confounded or the causal model provides the simplest graphical model. 
%%We achieve this by finding the simplest graphical model in terms of Kolmogorov complexity.
%%As Kolmogorov complexity is not directly computable, we employ the minimum description length to approximate it.
%Third, we investigate dataset harmonization techniques to \emph{remove} bias in imaging features, where we propose an extension of the recently introduced ComBat algorithm to control for global variation in image features.  
%We empirically show that harmonization can reduce dataset specific information in image features, but that it can also easily remove relevant subject-specific information. 
\end{abstract}


\end{frontmatter}

\section{Introduction}

Is it possible to predict the dataset that a brain scan is coming from based on image-derived measures like volume or thickness? 
Initially, we may guess that it should be impossible. 
On second thought, we may notice that it depends on the datasets we are comparing. %, although the type of datasets that we compare will effect the result. 
If one dataset only contains adolescent subjects and another one only \revision{elderly} subjects, it should be possible to distinguish them due to the association of image-derived measures with age. 
%If we would like to differentiate between one dataset with older subjects and one dataset with younger subjects, it should be possible because age is associated with image measurements. 
%But next to these subject specific characteristics that we would like to have in the derived features, are there unique signatures in each dataset that should allow for a correct identification, even if the demographics are similar?
But next to demographics, are there unique signatures in each dataset that would facilitate identifying the source of an image?
We will demonstrate that the source dataset can indeed be identified with much higher accuracy than would be expected from basic demographics of the participants.
Hence, not only subject-specific, but also imaging site-specific information is implicitly encoded.
This is insofar surprising as we are working with image-derived measures, which should only relate to the underlying neuroanatomy of the subject and not to the imaging site, where the scan was acquired.

Colloquially, this phenomenon is referred to as \emph{dataset bias}.
In statistics, bias refers to a \revision{systematic} deviation from
the true, possibly unknown, underlying quantitative parameter that is being estimated.
In a typical neuroimaging study, various types of bias can be present that can
alter the conclusions one deduces from this study.
In the first step, individuals have to be enrolled into the study.
If subjects do not faithfully represent the overall population one wants to
study, i.e., one obtains a non-random sample of a population, conclusions will be biased. This is referred to as \emph{selection bias}.
For instance, selection bias is present if the study recruits particular target groups, e.g., young adults or patients with a particular disease.
While such a selection may be related to the study objective and therefore done on purpose, there are also hidden, unintended factors like the over-representation of more educated participants~\citep{smith2018statistical} that would yield biased estimates with respect to the overall population.
%For instance, if some members of the population are less likely to participate or to be included because they did not obtain an academic degree, estimates would be biased with respect to the overall population~\citep{smith2018statistical}.

After enrollment, subjects will undergo magnetic resonance imaging.
Prior studies on inter-scanner variability have already noted that there is a strong dependence of the acquired image on magnetic field strength, manufacturer, gradients, pulse sequences, and head positioning~\citep{jovicich2009mri}.
%While each subject has only a single brain, we obtain different images
%of the same brain depending on magnetic field strength, manufacturer, gradients, pulse sequences and head positioning.
While standardization efforts are undertaken for instance by the ADNI~\citep{jack2008alzheimer}, variations related to the scanner remain~\citep{kruggel2010impact}, and it is even questionable if a further standardization is in the manufacturer's interest.
Even when assuming a faithful image reconstruction, scans often undergo
various image analysis steps to derive summary statistics that depend on the
algorithm being used.
The choice of segmentation and registration algorithms can cause varying results, also subject to the input, potentially affected by motion artifacts, voxel sizes, and image noise.
Therefore, neuroimaging data is unavoidably subject to various types of \emph{measurement biases}.

Once data has been collected and processed, usually
image-derived measures are related to disease status
or outcomes of a neurocognitive test to determine which
brain structures are responsible for the observed outcome.
It is important to remember that, in general, regression analysis can only
establish correlation, but not causation.
Association, unlike causation, is a symmetric relationship:
two variables $X$ and $Y$ are associated, regardless of whether
$X$ causes $Y$, or $Y$ causes $X$.
Only under exchangeability, correlation implies causation~\citep{hernan}.
A common violation of  exchangeability is the presence of confounding. 
%A prominent reason for exchangeability to not hold is the presence of confounding. 
For example, consider you want to study causes of Alzheimer's disease (AD). 
Analyses of the data show a high correlation between gray hair and AD, which may na\"ively lead to the conclusion that gray hair causes AD. 
However, the observed correlation between gray hair and AD
is only due to a person's age. 
Therefore, the association between gray hair and AD is confounded
by the common cause age.
This form of bias is known as \emph{confounding bias}.

%Consider a scenario where want to study how drinking coffee affects
%developing Alzheimer's disease (AD).
%Assume the genetic alteration APOE, which has a known
%causal effect on AD, is more frequent in patients that regularly drink coffee.
%A regression model would certainly be able to accurately predict AD status from the amount of coffee a subject drinks.
%However, the observed correlation between amount of coffee and AD
%is only due to an unmeasured genetic factor $U$ that is in linkage disequilibrium
%with the APOE alteration and also increases subjects' craving for coffee.
%Therefore, the association between drinking coffee and AD is confounded
%by a common cause $U$.
%This form of bias is known as \emph{confounding bias}.

Dataset bias is becoming pivotal as neuroimaging is joining the ranks of a ``big data'' science with more and larger datasets becoming available~\citep{smith2018statistical}.
Large sample sizes are required for a number of applications in
neuroimaging, such as association studies in imaging-genetics,
or training complex machine learning models -- in particular in deep learning.
As outlined above, sources of bias are plentiful and depend on
the research question and the data that is used to answer it.
For this reason, neuroimaging data is usually collected with a particular research question in mind, and inclusion criteria are tailored to answering this particular question as unbiased as possible, e.g., by randomization or collecting information
on possible confounders.
Therefore, pooling data from studies that have been designed with different research questions in mind, will likely lead to bias in a machine learning model trained
on this data.
In contrast, if a model would be truly unbiased on a population level, it would naturally generalize to other datasets.

\begin{table*}[t]
  \caption{\label{tab:dataStats}%
    Overview of neuroimaging datasets used in this study.}%
  \fontsize{8}{8}\selectfont%
  \renewcommand{\arraystretch}{1.3}%
  \rowcolors{2}{}{gray!20}%
  \center
  \begin{tabular}{llrrrrrr}
  \toprule
Dataset	&	Diagnosis	&	  $N$	& \ \ \ Age (mean)  	& \ \ 	Age (SD)	& \ \ 	Males \%	 & \ \ 	Sites	& \ \	Diseased	\\
\midrule
ABCD	&		&	8,751	&	9.9	&	0.6	&	51.6	&	29	&	-	\\
ABCD Subset	&		&	(1,570)	&	9.9	&	0.6	&	51.6	&	29	&	-	\\
ABIDE I	&	Autism	&	\ \ \ \ \ 1,095	&	17.1	&	8.1	&	85.2	&	24	&	526	\\
ABIDE II	&	Autism	&	1,032	&	15.2	&	9.4	&	76.1	&	17	&	477	\\
ADHD200	&	ADHD	&	965	&	12.1	&	3.3	&	61.8	&	8	&	384	\\
ADNI	&	Alzheimer's	&	1,682	&	73.6	&	7.2	&	55.0	&	62	&	1,144	\\
AIBL	&	Alzheimer's	&	262	&	72.9	&	7.6	&	47.3	&	2	&	91	\\
COBRE	&	Schizophrenia	&	146	&	37.0	&	12.8	&	74.7	&	1	&	72	\\
CORR	&		&	1,476	&	25.9	&	15.4	&	50.0	&	32	&	0	\\
GSP	    &		&	1,563	&	21.5	&	2.8	&	42.3	&	5	&	0	\\
HBN	    &	Psychiatric	&	689	&	10.7	&	3.6	&	59.7	&	2	&	497	\\
HCP	    &		&	1,113	&	28.8	&	3.7	&	45.6	&	1	&	0	\\
IXI	    &		&	561	&	48.6	&	16.5	&	44.6	&	3	&	0	\\
MCIC	&	Schizophrenia	&	194	&	33.1	&	11.5	&	71.6	&	3	&	104	\\
NKI	    &	Psychiatric	&	624	&	38.4	&	22.5	&	39.1	&	1	&	268	\\
OASIS	&	Alzheimer's	&	415	&	52.8	&	25.1	&	38.6	&	1	&	100	\\
PPMI	&	Parkinson's	&	390	&	61.2	&	10.0	&	62.6	&	16	&	284	\\
UKB	&	Diverse 	&	14,362	&	62.8	&	7.5	&	47.6	&	2	&	1,878	\\
UKB Subset	&	Diverse 	&	(1,474)	&	63.2	&	7.4	&	47.4	&	2	&	160	\\
\midrule
& & 35,320 & 39.6 & 25.9 & 51.6 & 210 & \\
  \bottomrule
 \end{tabular}
\end{table*}	

\begin{figure}[t]
\begin{center}
	\includegraphics[width=0.49\textwidth]{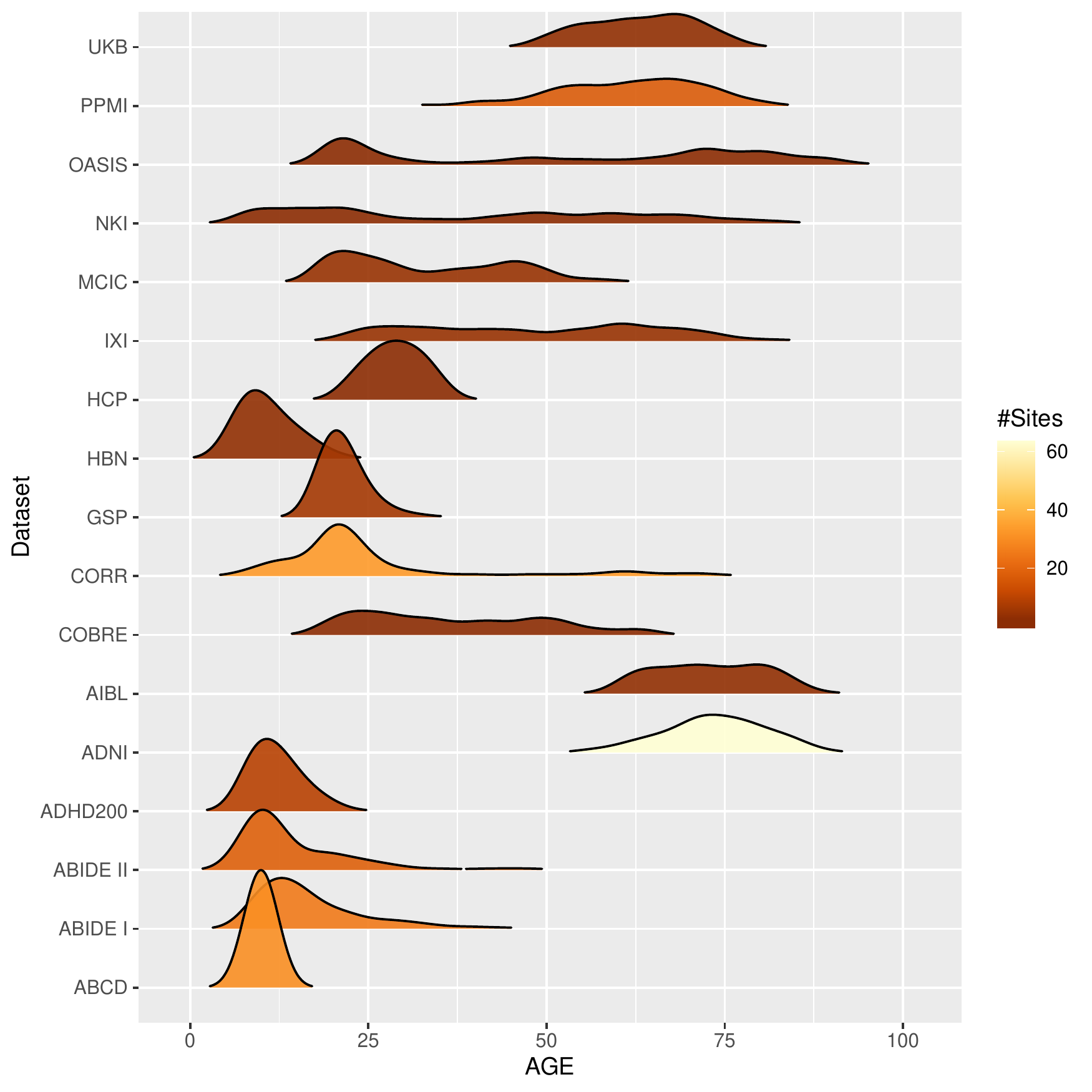} 
\caption{Age distribution per dataset. The fill color corresponds to the number of imaging sites per dataset.
%\todo[inline]{ADNI domoniates. Should be discretize the \#sites color scale (1, 2-5, 5-10, 10-30, 30+)?}
\label{fig:ageDist}%
}
\end{center}
\end{figure}

%\subsection{Outline}
In this article, we study bias in neuroimaging data. To this end, we combine data from 17 large-scale studies, presented in section~\ref{sec:neuroimaging-data}.
First, we propose \emph{Name That Dataset}, i.e., the prediction of the dataset that a scan is part of, as an experiment to detect inter-dataset bias in section~\ref{sec:nameThat}.
Second, we take a closer look at confounding bias in section~\ref{sec:causal} and present a method for distinguishing between causal and confounded statistical relationships in a single dataset using causal inference. 
As we do not know all potential confounders, we work with a model that assumes an unobserved confounding variable. 
%To this end, we use the algorithmic Markov condition to determine the simplest model in terms of Kolmogorov complexity to determine the true causal model. 
\revision{Third, we discuss methods for dataset harmonization as a means of reducing bias in the data, and present an extension that accounts for global variation across features  in section~\ref{sec:harmonization}.} %\todo[inline]{I don't like ``bias in the data'', but couldn't come up with something better.}
Finally, we present results for dataset prediction on harmonized data, brain age prediction with harmonization, and the impact of harmonization on two causal models in section~\ref{sec:results}. 
%discuss, extend, and evaluate methods for dataset harmonization as a means of reducing bias in the data. 
An earlier version of this work has been presented at a conference~\citep{wachinger2019quantifying} and is extended in this article with more datasets, details, experiments, and the addition of harmonization.

\section{Neuroimaging Data}
\label{sec:neuroimaging-data}

We work on MRI T1 brain scans from 17 large-scale public datasets: 
\begin{enumerate}
\item Adolescent Brain Cognitive Development (ABCD)~\citep{casey2018adolescent},
\item Autism Brain Imaging Data Exchange (ABIDE)~I~\citep{ABIDE},
\item \revision{Autism Brain Imaging Data Exchange (ABIDE)~II~\citep{di2017enhancing}},
\item Attention Deficit Hyperactivity Disorder (ADHD200)~\citep{ADHD200},
\item Alzheimer's Disease Neurimaging Initiatie (ADNI)~\citep{jack2008alzheimer} (for up-to-date information, see \url{www.adni-info.org}),
\item Australian Imaging Biomarkers and Lifestyle Study of Ageing (AIBL)~\citep{AIBL},
\item Center for Biomedical Research Excellence (COBRE)~\citep{COBRE},
\item Consortium for Reliability and Reproducibility (CORR)~\citep{zuo2014open},
\item Genomic Superstruct Project (GSP)~\citep{GSP},
\item Healthy Brain Network (HBN)~\citep{alexander2017open},
\item Human Connectome Project (HCP)~\citep{HCP},
\item IXI Dataset,\footnote{\url{http://brain-development.org/ixi-dataset/}}
\item Open Access Series of Imaging Studies (OASIS) cross-sectional sample~\citep{OASIS},
\item MIND Clinical Imaging Consortium (MCIC) schizophrenia sample~~\citep{MCIC},
\item Nathan Kline Institute -- Rockland Sample (NKI)~\citep{nooner2012nki},
\item Parkinson Progression Marker Initiative (PPMI)~\citep{PPMI},
\item UK Biobank Imaging (UKB)~\citep{miller2016multimodal}.
\end{enumerate}

All datasets were processed with FreeSurfer~\citep{fischl2002whole} version 5.3. 
We keep only one scan per subject from longitudinal or test-retest datasets. 
%After exclusion of scans with processing errors and incomplete meta data,
%scans from 12,207 subjects (6,827 male; 8,126 controls; mean age: $33.5 \pm 23.9$) remained.
After exclusion of scans with processing errors and incomplete meta-data,
scans from 35,320 subjects remained. 
Fig.~\ref{fig:ageDist} illustrates the age distribution per dataset together with the number of sites. 
Table~\ref{tab:dataStats} presents an overview of demographics per dataset. 
The ABCD and UKB datasets stand out as the largest datasets, which introduces a \revision{severe} class imbalance for the dataset prediction experiment in section~\ref{sec:nameThat}. %\todo[inline]{The experiment has been mentioned yet.}
Hence, we downsampled ACBD and UKB data to obtain a sample size close to the next largest datasets. 
We preserved the heterogeneity of the datasets by ensuring that all imaging sites are included in the subset. 
For UKB, we marked subjects with psychiatric diseases and cancer in the head as diseased.
For ABCD, the diagnostic information is highly complex as mental health of adolescents is investigated in general, so that we have not encoded it for this study. 
%As disease is not in the focus of this study, we have not encoded diagnosis for ABCD. 

\section{Name That Dataset} 
\label{sec:nameThat}

\begin{figure*}[t]
\begin{center}
	\includegraphics[width=0.47\textwidth]{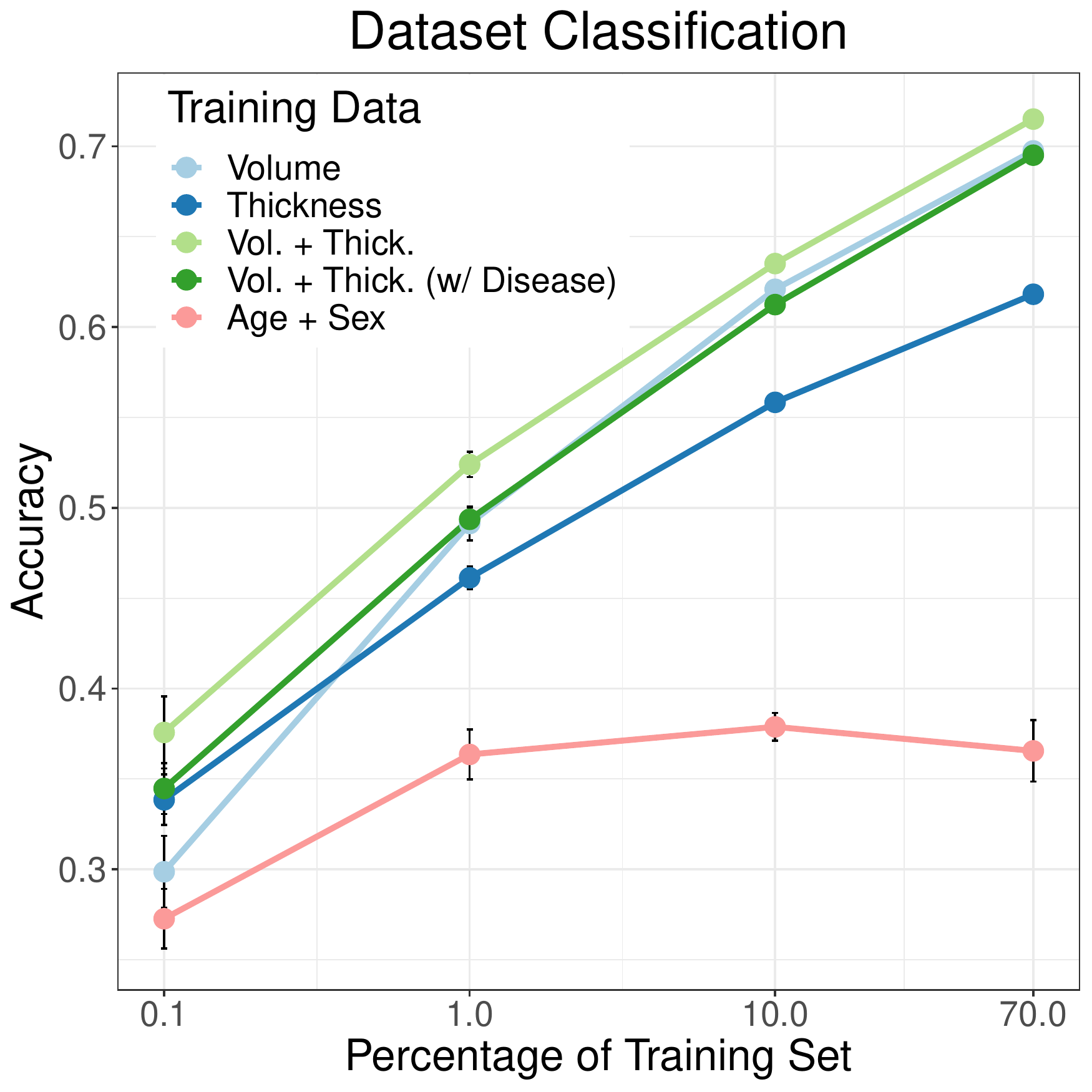} 
	\includegraphics[width=0.52\textwidth]{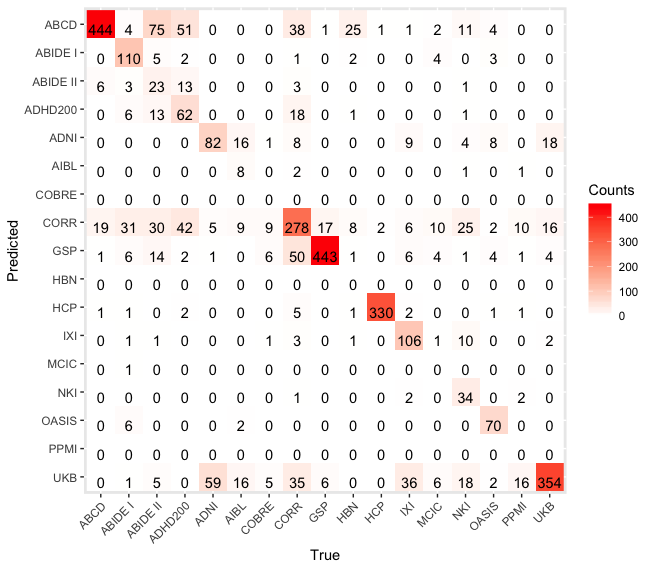} 
\caption{Left: Dataset classification accuracy for age and sex, volume, thickness, and their combination. The percentage of the data used for training is shown in log-scale. Lines show the average score over 50 repetitions, error bars show the standard deviation.
Right: Confusion matrix for volume and thickness with 70\% training data. %
\label{fig:DatasetClass}%
%\todo[inline]{Left: Can we use another color palette that's not paired for volume and thickness?}
}
\end{center}
\end{figure*}

In order to detect the impact of dataset bias, we  play the game \emph{Name That Dataset} on neuroimaging data that was originally proposed by~\cite{torralba2011unbiased} on natural images. 
The task is to predict the dataset~$D$ an MRI scan is coming from solely based on \revision{image-derived measures~$\Psi$}, where we use volume and thickness.\footnote{\revision{We used 55 volume measures from FreeSurfer as listed in aseg.stats and 70 mean thickness measures of cortical parcels as listed in lh.aparc.stats and rh.aparc.stats.}}
A random forest classifier is trained for the \revision{prediction $\Psi \mapsto D$}, where we use default settings from the R package by~\cite{liaw2002classification}.\footnote{500 trees, mtry is square root of number of variables}
If dataset bias would be absent, we would expect a prediction accuracy
close to random chance, which would be 5.9\% for 17 datasets; taking unequal dataset sizes into account, it would be 7.8\%.
As additional reference, we report the results of a classifier trained on age and sex alone, which can achieve higher accuracy than random chance as datasets focus on specific parts of the population. 
%In the experiments, we split data into training and testing sets with stratified sampling by dataset to ensure that each dataset is accordingly represented. 
\revision{In the experiments, we split data into training and testing sets with stratified sampling by dataset. This ensures that also the small datasets are represented in the training set for low sampling rates.}

Fig.~\ref{fig:DatasetClass} illustrates the performance for classifying images into one of 17 datasets from volume and thickness measures, and their combination. 
We mainly report results on healthy controls  to exclude disease-specific effects that could potentially ease classification; for the combination of volume and thickness we also show results for including diseased subjects, indicated as `(w/ Disease)'. 
In this experiment, we vary the amount of training data from 0.1\% to 70\%. 
%As not all datasets have the same size and have different distributions of age and sex, we compare to results of a classifier trained on age and sex alone as baseline. 
The classification accuracy widely improves as the training set increases, particularly for the image features. 
The highest accuracy of 71.5\% is achieved by the combination of volume and thickness measures, which perform better than each of them alone. 
Except for the 0.1\% training set size, volumes have a higher accuracy than thickness measures; for 70\% training, the difference in accuracy is 7.9\%. 
The accuracy after the inclusion of diseased patients is slightly lower compared to healthy subjects, as shown for the combination of volume and thickness. 
%With only 0.1\% of the data used for training, volume measures perform similar to prediction with age and sex. 
%As we increase the amount of training data to 70\%, the accuracy increases to 73.3\% for the combination of volume and thickness features, which perform better than each of them alone. 
The classifier with age and sex reaches 37.9\% accuracy, which is well above random and therefore hints at selection bias. 
Yet, compared to over 70\% accuracy for image features, there must be another source of bias that cannot be explained by basic demographics, such as measurement and confounding bias.

From the confusion matrix in Fig.~\ref{fig:DatasetClass}, the high classification accuracy (diagonal elements) indicates that datasets possess unique, identifiable characteristics.
In addition, we can see that datasets with a similar population
%\todo[inline]{How about ordering rows by age range, e.g. using hierarchical clustering?}
result in higher confusion, e.g., between ABCD, ABIDE I+II, and ADHD200. 
Single-site datasets, like HCP, that have strict inclusion criteria and
imaging protocols show almost no confusion with any of the other datasets.
In contrast, multi-site datasets like CORR that also cover a wide age range, show high confusion with other datasets. 
Also for UKB, with its large size and age distribution, we observe confusion with other datasets, although scans were only acquired at two sites with the same scanner.
% \todo[inline]{I think this should be the first point to discuss.}

The lesson learned from this experiment is that even when working with image-derived values that represent physical measures (volume, thickness), substantial bias in datasets remains, despite techniques like atlas renormalization~\citep{han2007atlas} were employed to improve consistency across scanners. 
Of course, much of the bias can be attributed to the different goals of the studies, like the inclusion of subjects from different age groups. 
However, even when focusing on datasets that cover a similar age range, we observe a high accuracy, e.g., ABIDE I and II. 
While we are not aware of previous attempts to \emph{Name That Dataset}, our results echo concerns raised in previous studies. 
In an ENIGMA study with over 15,000 subjects on brain asymmetry \citep{guadalupe2017human}, it was reported that dataset heterogeneity explained over 10\% of the total observed variance per structure. 
In ADNI, which has an optimized MPRAGE imaging protocol across all sites~\citep{jack2008alzheimer}, the intra-subject variability of compartment volumes for scans on different scanners was roughly 10 times higher than repeated scans on the same scanner~\citep{kruggel2010impact}.  
Similarly, \cite{wachinger2016domain} reported a drop in accuracy for Alzheimer's disease prediction when training and testing on different datasets.

\section{Telling Causal from Confounded with Causal Inference}
\label{sec:causal}

%In causal inference, a \emph{systematic bias} is present when the data are insufficient to compute the causal effect even with an infinite sample size.
%Commonly, this arises from a structural association between treatment and outcome that does not arise from the causal effect of treatment on outcome in the population of interest. 
%Confounding bias, when treatment and outcome share a common cause. 
%Selection bias, when conditioning on common effects of treatment and outcome. 
%Measurement bias, when measurement errors are present. 

\begin{figure}[t]
    \centering%
    \includegraphics[width=0.48\textwidth]{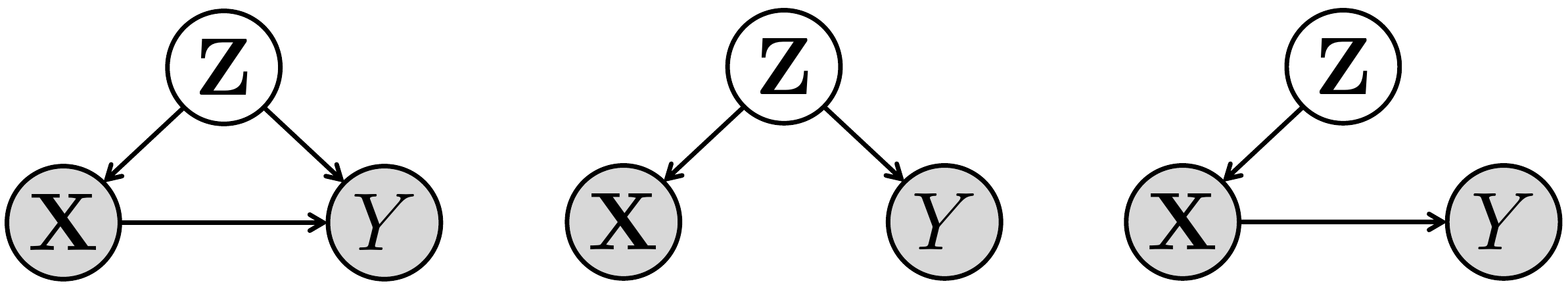}%
    \caption{Probabilistic graphical models for observed variables $\bbX$, $Y$ and unobserved confounders $\bbZ$. The statistical relationship between $\bbX$ and $Y$ is due to confounder~$\bbZ$ and due to the influence of $\bbX$ on $Y$ (left). Limiting cases are pure confounding (middle) and pure causality (right).%
    \label{fig:causalDiag}%
    }%
\end{figure}

In the previous section, we have established that there is correlation between a
feature vector \revision{$\psi$}, derived from MRI scans, and the dataset $D$ the scan belongs to,
by estimating the probability \revision{$P(D=d\,|\,\Psi = \psi)$} via a random forest classifier.
While this has yielded useful insights, it only provides a measure of statistical dependence,
which alone is insufficient to determine causal structures of confounding bias.

\subsection{Confounding Bias in Causal Inference}
Here, we want to study bias in a more \revision{principled}  manner by
looking at confounding bias in a causal inference framework. 
Generally, it is challenging to infer cause and effect from observational data, as this normally requires randomized controlled trials~\citep{Pearl2000}. 
The problem is that it is difficult to determine whether a variable $X$ causes $Y$ or whether both variables have a common cause $Z$. %\todo[inline]{Define X,Y,Z.}
If $X$ and $Y$ share the cause $Z$, then confounding is present and $Z$ is referred to as confounder, as illustrated in Fig.~\ref{fig:causalDiag}. 
Because there is an open backdoor path between $X$ and $Y$, it is unclear if an association between $X$ and $Y$ results from the causal effect $X \rightarrow Y$ or from the path with the common cause $X \leftarrow Z \rightarrow Y$. 
The extreme cases would be the purely causal setting without confounding $X \rightarrow Y$, illustrated in Fig.~\ref{fig:causalDiag} (right) and the purely confounded setting, where the correlation between $X$ and $Y$ is entirely due to common cause $Z$, illustrated in Fig.~\ref{fig:causalDiag} (middle).

For inferring causality, one of the core assumptions is \emph{causal sufficiency}, i.e., we know all potential confounders. 
This assumption is often violated in practice, since we do not know all potentially relevant factors nor do we have data on them, yielding spurious inferences. 
Even if they would be known, they would not be available in all publicly available neuroimaging datasets. 
Consequently, we consider a recently proposed approach for causal inference by~\cite{kaltenpoth2019we} that assumes an \emph{unobserved} confounding variable. 
The method explicitly models the hidden confounder with probabilistic PCA, which allows comparing the causal  $X \rightarrow Y$ and confounded model $X \leftarrow Z \rightarrow Y$ to conclude whether the relationship is causal or confounded. 
Next, we will present the details of the approach. 
\subsection{Assumptions for Inferring Causality with Unknown Confounders}

We consider samples from the joint distribution $P(\bbX,Y)$ over two statistically dependent continuous-valued random variables, where $\bbX$ can be multivariate  and $Y$ is univariate.
The aim is to determine whether it is more likely that $\bbX$ causes $Y$, or that  there exists an unobserved random variable $\bbZ$ that is the cause of both $\bbX$ and $Y$. 
$\bbZ$ can be multivariate. 
Inferring causal relations from observational data is only attainable under specific model assumptions.
A general assumption in causal inference is that the data distribution can be represented as a
causal directed acyclic graph that satisfies the Markov factorization property:
conditioned on its parents (direct causes), each variable is independent of its nondescendants
~\citep{Pearl2000}.
In addition, faithfulness and the previously mentioned assumption of causal sufficiency are required.
Faithfulness implies that if $\bbX$ is independent of $Y$, there is no direct influence
between the two in the underlying graph~\citep{Pearl2000}. %\todo[inline]{Actually, we would need to talk about d-seperation to explain Faithfulness.}
While faithfulness is a strong condition, it is generally reasonable.
Establishing causal sufficiency is more problematic in practice.
%\todo[inline]{I find this part confusing, because causal sufficiency was defined in the previous section, then forgotten, and picked up again, while the author assumptions are all defined here.}

Our approach incorporates the algorithmic Markov condition (AMC)~\citep{janzing2010causal}, which states that if $\bbX$ causes $Y$, the factorization of the joint distribution $P(\bbX,Y)$ in the true causal direction has a lower Kolmogorov complexity than in the anti-causal direction. 
Together with causal sufficiency, this allows to identify the true causal network as the least complex one. 
Although the AMC generally relies on causal sufficiency,  \cite{kaltenpoth2019we} proposed to \revision{incorporate} the confounder as a latent variable $\bbZ$. 
This approach allows for computing whether confounding is present, without explicitly knowing the confounder or having data on it, which is  very relevant in practice. 

A remaining challenge for estimating the AMC is the evaluation of the Kolmogorov complexity as it is not directly computable. 
To this end, the minimum description length (MDL) principle provides a statistically well-founded approach to approximate Kolmogorov complexity~\citep{grunwald2007minimum}. 
Considering model class $\mM$ and a fully Bayesian formulation, the code length function~$L$ is computed as 
%\todo[inline]{something seems to be missing to conclude this sentence.}
\begin{equation}
L(\bbX | \mM) = - \log \int_{M \in \mM} P(\bbX | M)  \mathrm{d} P(M),
\end{equation}
where $P(M)$ is a prior on the model class $\mM$.

%Inferring causal relations from observational data is challenging and is only attainable under specific model assumptions.
%We use an approach that incorporates the algorithmic Markov condition~\cite{kaltenpoth2019we},
%which states that the simplest factorization of the joint distribution $P(X, Y)$, in terms of
%Kolmogorov complexity, corresponds to the true generative process.
%As Kolmogorov complexity is not directly computable, we employ the minimum description length to approximate it.

\subsection{Causal and Confounded Models}

Now that we established the theoretical foundation of our work, we need to define
two factorizations of $P(\bbX, Y)$ -- one for the causal and one for the confounded scenario --
to decide whether the relationship between $\bbX$ and $Y$ is indeed causal or not.
Figure~\ref{fig:causalDiag} illustrates the two factorizations we consider here:
\begin{enumerate}
\item Causal: $P(\bbX, Y) = P(Y| \bbX) P(\bbX|\bbZ) P(\bbZ)$, which is represented by a linear regression
model,
\item Confounded: $P(\bbX, Y) = P(Y|\bbZ) P(\bbX|\bbZ) P(\bbZ)$, which is estimated by probabilistic PCA~\citep{Tipping1999}.
\end{enumerate}
%(i) the causal model $P(X, Y) = P(Y|X) P(X|Z) P(Z)$ is represented by a linear regression
%model, and (ii) the confounded model $P(X, Y) = P(Y|Z) P(X|Z) P(Z)$ by probabilistic PCA. 
%\revision{To emphasize that $X$ and $Z$ can be multivariate, we continue with $\bbX$ and $\bbZ$, whereas $Y$ is required to be univariate.} 
The complexity under the causal model can be estimated by minimum description length
$L_{\text{ca}}(\bbX, Y)$ via
\revision{
\begin{equation}
\begin{split}%
    \label{eq:complexity_causal}%
    L_{\text{ca}}(\bbX, Y) &= - \log P(\bbX) \int P(Y | \bbX, \bbw) P(\bbw)  \mathrm{d} \bbw , \\
    \bbX &\sim \mN(\mathbf{0}, \sigma_x^2I), \\
    \bbw &\sim \mN(\mathbf{0},\sigma_w^2I), \\
    Y | \bbX, \bbw &\sim \mN(\bbw^\top \bbX,\sigma_y^2I) .
\end{split}
\end{equation}}
The complexity of the confounded model is estimated by
\revision{
\begin{equation}
  \begin{split}%
      \label{eq:complexity_confounded}%
      L_{\text{co}}(\bbX,Y) &= \\ - &\log  \int P(\bbX,Y | \bbZ, \bbW)  P(\bbZ) P(\bbW)  \mathrm{d} \bbW \mathrm{d} \bbZ, \\
      \bbZ &\sim \mN(\mathbf{0}, \sigma_z^2I), \\
      [\bbW, \mathbf{w}_y] &\sim \mN(\mathbf{0},\sigma_w^2I), \\
      \bbX | \bbZ, \bbW &\sim \mN(\bbW^\top \bbZ,\sigma_x^2I), \\
      Y | \bbZ, \bbW &\sim \mN(\mathbf{w}_y^\top \bbZ, \sigma_y^2) ,
  \end{split}
  \end{equation}
  where the confounders $\bbZ$ and the principal axes $\bbW$, $\mathbf{w}_y$ are estimated using probabilistic PCA by appending $Y$ as an extra column
  to $\bbX$}~\citep{Tipping1999}.

Note that we do not require that the confounders are known or measured; since
we marginalize over $\bbZ$, we only need to specify its dimensionality $k$.
To compare the causal ($\bbX \rightarrow Y$) and the confounded model
($\bbX \leftarrow \bbZ \rightarrow Y$),
we  compute 
\begin{equation}
\Delta(\bbX, Y) = L_{\text{co}}(\bbX, Y) - L_{\text{ca}}(\bbX, Y).
\end{equation}
If the causal model better describes the data than the confounded model,
we obtain $\Delta(\bbX, Y) > 0$ -- the more positive, the more confident we are.
If instead $\Delta(\bbX, Y) < 0$, the roles are reversed.
As $\Delta$ is dependent on the size of the dataset, we also consider the normalized version $\frac{1}{N} \Delta$, with $N$ denoting the number of subjects.

% We estimate the complexity of both models with automatic differentiation variational inference~\citep{kucukelbir2017automatic}, efficiently implemented in PyMC3, and use $k=1$ as the dimensionality of $\bbZ$ throughout our experiments. 
\revision{We first estimate the posterior distribution of the model parameters
$\sigma_x, \sigma_y, \mathbf{w}_y, \bbW, \bbZ$
via Markov chain Monte Carlo using a no-U-turn sampler~\citep{Hoffman2014}
with four chains and 8,000 samples each (of which 1,500 are used for burn-in)
as implemented in RStan~2.19.3~\citep{Carpenter2017,RStan}.
We experienced instability estimating the posterior using variational inference as proposed by~\cite{kaltenpoth2019we}.
%Note that this is different to the implementation with PyMC3 by~\cite{kaltenpoth2019we}, for which experienced instability.
The integral over the model parameters in the marginal likelihood in Eqs.~\eqref{eq:complexity_causal}
and \eqref{eq:complexity_confounded}
is analytically intractable and we need to approximate it via
Monte Carlo sampling.
Instead of using the na{\"i}ve Monte Carlo estimator of the marginal
likelihood, we estimate it via bridge sampling, which has lower
variance~\citep{FruehwirthSchnatter2004,bridgesampling}.
We repeated this process ten times to obtain an interval
of possible values for $L_{\text{ca}}$ and $L_{\text{co}}$.}

\section{Data Harmonization}
\label{sec:harmonization}

%\revision{TODO: motivate with Blei paper?}

In the previous sections, we described methods for identifying confounding bias.
\revision{Here, we want to study whether methods for harmonizing neuroimaging datasets can effectively reduce bias, and therefore can enable to identify causal effects.} 
Harmonization of multi-site data aims to remove unwanted variability associated with scanner and site when
pooling data.
We refer to such unwanted variation as \emph{non-biological variation} to distinguish it from \emph{biological variation} about the subject, which we want to retain. 
\revision{Current methods for harmonization focus on adjusting for known confounders, but as argued in section~\ref{sec:causal}, neither do we know all confounding factors nor do we have data on them.
We avoid this issue by introducing a novel harmonization approach that can be used
in the presence of unknown confounders by computing substitute confounders.}
%to discuss existing methods for harmonizing neuroimaging datasets and study wether they are effective in reducing bias. 
%\todo[inline]{First define what Harmonization is?}

Harmonization can be performed at multiple stages: by specifying a strict acquisition protocol,
by normalization in image processing, or by adjusting image-derived features before data analysis. 
Here, we will focus on harmonization at the feature stage, as we cannot influence the acquisition anymore, and we are using FreeSurfer for the image analysis, which already has a sophisticated normalization pipeline. 

We will discuss several existing methods for harmonization,
\revision{and propose our novel extension to account for unknown confounders below}.
Table~\ref{tab:models} summarizes the different harmonization models and the corresponding variable updates. 
First, we need to specify on which level we want to harmonize features. 
For multi-site datasets, the obvious choice is to harmonize scans with respect to imaging site. 
Based on your initial motivation for \emph{Name That Dataset}, an alternative would be to harmonize with respect to image source (dataset). 
We will evaluate both in our experiments, but focus on imaging site in the following description. 

%At the acquisition stage, the variability across imaging sites is reduced by standardizing the imaging protocol. To this end, MRI sequences like the ADNI protocol have been developed to acquire similar scans across manufacturers. 
%Other strategies include the usage of the same scanner model at different centers. 
%For the image processing to obtain the segmentation of the scan, techniques have been developed to normalize scans to ensure a similar output. 

\begin{table*}
\centering
\small
%\begin{tabular}{ c c || c | c }
\renewcommand{\arraystretch}{1.5}
\begin{tabular}{ l r r@{\,=\,}l r@{\,=\,}l }
\toprule
 Variables&& \multicolumn{2}{c}{Model} & \multicolumn{2}{c}{Update} \\
 \midrule
\multirow{2}{*}{Site}
 & Linear &
   \(\displaystyle Y_{ijf}\) & \(\displaystyle \alpha_f + \gamma_{if}  + \varepsilon_{ijf} \) &
   \(\displaystyle \mY_{ijf}\) & \(\displaystyle Y_{ijf} -  \hat{\gamma}_{if} \)  \\
 & ComBat &
   \(\displaystyle Y_{ijf}\) & \(\displaystyle \alpha_f + \gamma_{if}  + \delta_i \varepsilon_{ijf} \) &
   \(\displaystyle \mY_{ijf}\) & \(\displaystyle \frac{Y_{ijf} - \hat{\alpha}_f -  \hat{\gamma}_{if}}{\hat{\delta}_i} + \hat{\alpha}_f \)  \\
\midrule
\multirow{2}{*}{Site \& Keep}
 & Linear &
   \(\displaystyle Y_{ijf}\) & \(\displaystyle \alpha_f + \gamma_{if}+ \bbeta_{f}^\top \bbk_j  + \varepsilon_{ijf} \)
 & \(\displaystyle \mY_{ijf}\) & \(\displaystyle Y_{ijf} -  \hat{\gamma}_{if} \)  \\
 & ComBat &
   \(\displaystyle Y_{ijf}\) & \(\displaystyle \alpha_f + \gamma_{if}  + \bbeta_{f}^\top \bbk_j + \delta_{i} \varepsilon_{ijf} \)
 & \(\displaystyle \mY_{ijf}\) & \(\displaystyle \frac{Y_{ijf} - \hat{\alpha_f} - \hat{\bbeta}_{f}^\top \bbk_j  - \hat{\gamma}_{if} }{\hat{\delta}_{if}} + \hat{\alpha}_f  + \hat{\bbeta}_{f}^\top \bbk_j \)  \\
\midrule
\multirow{2}{2cm}{Site \& Keep \& Remove}
 & Linear &
   \(\displaystyle Y_{ijf}\) & \(\displaystyle \alpha_f + \gamma_{if}  + \bbeta_{f}^\top \bbk_j + \bzeta_{f}^\top \bbr_j +  \varepsilon_{ijf} \)
 & \(\displaystyle \mY_{ijf}\) & \(\displaystyle Y_{ijf} -  \hat{\gamma}_{if} - \hat{\bzeta}_{f}^\top \bbr_j\)  \\
 & ComBat &
   \(\displaystyle Y_{ijf}\) & \(\displaystyle \alpha_f + \gamma_{if}  + \bbeta_{f}^\top \bbk_j + \bzeta_{f}^\top \bbr_j +  \delta_{i} \varepsilon_{ijf} \)
 & \(\displaystyle \mY_{ijf}\) & \(\displaystyle \frac{Y_{ijf} - \hat{\alpha}_f - \hat{\bbeta}_{f}^\top \bbk_j - \hat{\bzeta}_{f}^\top \bbr_j - \hat{\gamma}_{if} }{\hat{\delta}_{if}} + \hat{\alpha}_f  + \hat{\bbeta}_{f}^\top \bbk_j
\) \\
\bottomrule
\end{tabular}
\caption{\label{tab:models}%
Summary of harmonization models for image features~$Y$ with the corresponding update equations for obtaining the harmonized value~$\mY$. 
The indices are imaging site~$i$, subject~$j$, and feature type~$f$. 
We consider linear regression and ComBat, together with different types for variables. Site is represented by~$\gamma$. 
For biological variation that we would like to keep~$\bbk$, we consider age and sex. 
For non-biological variation that we would like to remove~$\bbr$, we consider manufacturer, field strength, and principal components. Estimated model parameters are denoted with hat.}
\end{table*}

\subsection{Harmonization only based on imaging sites} %\todo[inline]{The section title doesn't fit well.}
A na{\"i}ve approach for harmonizing across imaging sites is \textbf{Z-score normalization}.  
Consider an image-derived measurement~$Y_{ijf}$ for imaging site~$i$, subject~$j$, and feature type~$f$. 
After computing the mean~$\hat{\mu}_{if}$ and standard deviation~$\hat{\sigma}_{if}$ for each imaging site and feature,
the feature's harmonized value $\mY_{ijf}$ is
\begin{equation}
\mY_{ijf} = \frac{Y_{ijf} -  \hat{\mu}_{if}} {\hat{\sigma}_{if}},
\end{equation}
where subject $j$ was scanned at site $i$. 
Z-score normalization is a standard procedure in statistics, simple to compute, but might remove information
that is required for downstream tasks, such as disease prediction.
Consider one imaging site with only young subjects and another one with only \revision{elderly} subjects,
both sites will have the same mean after the normalization, which for structures like the ventricles is inadequate~\citep{scahill2003longitudinal}.

As second harmonization approach controls for site-specific effects by computing residuals.
To this end, site becomes a regressor in a \textbf{linear regression} and the
feature's harmonized value is the residual: %\todo[inline]{Should we cite something here?}
% A separate regression model is fitted for each feature $f$
\begin{align}
Y_{ijf} &= \alpha_f + \gamma_{if} + \varepsilon_{ijf}, \\
\mY_{ijf} &= Y_{ijf} -  \hat{\gamma}_{if},
\end{align}
%\todo[inline]{$Y$ has now to meanings, the actual value, and the outcome variable in the linear model, maybe
%write $Y_{ijf} \sim \mathcal{N}(\alpha_f + \gamma_{if}, \sigma^2)$.}
where $\alpha_f$ is the average measure for the reference site, $\gamma_{if}$ is an additive imaging site effect, and $\varepsilon_{ijf}$ is the variance.
The parameters can be estimated by solving the corresponding
ordinary least squares problem.

%$Y_i$ is the image feature, $X_i$ is either the dataset or the imaging site, $A_i$ is the age, $S_i$ is the sex. 

\subsection{ComBat Harmonization} 
As an extension of the previous model, \citet{fortin2018harmonization} proposed harmonization of cortical thickness measures based on Combining Batches (ComBat)~\citep{johnson2007adjusting} from gene expression analysis. 
ComBat adds a site-specific scaling factor $\delta$, yielding a model that adjusts for additive and multiplicative effects. 
In addition, ComBat uses empirical Bayes for inferring model parameters, which assumes that model parameters across features are drawn from the same distribution. 
Hence, model parameters are not estimated independently per feature anymore, which can be helpful with small sample sizes, but also assumes that parameters across features are homogeneous. 

Accounting for imaging site can help in removing unwanted variation in the data, but at the same time it may have the detrimental effect of removing relevant information. 
For instance, we would like to keep subject-specific information due to biological variability. 
We add the vector~$\bbk$ to the model to indicate variables, whose influence we would like to keep after harmonization. 
\revision{In our experiments, we use sex and a linear and quadratic age term to account for non-linear aging effects~\citep{walhovd2011consistent}.} 
The model is 
\begin{equation}
Y_{ijf} = \alpha_f + \gamma_{if}  + \bbeta_{f}^\top \bbk_j +  \delta_{if} \varepsilon_{ijf}
\end{equation}
with the multiplicative imaging site effect~$\delta_{if}$ and vector of regression coefficients~$\bbeta_f$. 
The harmonized value is computed as
\begin{equation}
\mY_{ijf} = \frac{Y_{ijf} - \hat{\alpha}_f - \hat{\bbeta}_{f}^\top \bbk_j - \hat{\gamma}_{if} }{\hat{\delta}_{if}} + \hat{\alpha}_f  + \hat{\bbeta}_{f}^\top \bbk_j.
\end{equation}
The linear model can also be adapted to accommodate variables to keep $\bbk$, as shown in Table~\ref{tab:models}.

\subsection{Harmonization for non-biological variability with ComBat} 
%\subsubsection{ComBat Harmonization with known and unknown sources of variability} 
As a novel extension, we propose to add the vector~$\bbr$ to the model that contains variables that relate to non-biological variability, which we would like to explicitly remove from the features. The updated ComBat model\footnote{The extension is available as ComBat++ in our repository \url{https://github.com/ai-med/Dataset-Bias}.} is  
\begin{equation}
Y_{ijf} = \alpha_f + \gamma_{if}  + \bbeta_{f}^\top \bbk_j + \bzeta_{f}^\top \bbr_j +  \delta_{if} \varepsilon_{ijf},
\end{equation}
with the vector of regression coefficients~$\bzeta_{f}$. 
The harmonized values are 
\begin{equation}
\mY_{ijf} = \frac{Y_{ijf} - \hat{\alpha}_f - \hat{\bbeta}_{f}^\top \bbk_j - \hat{\bzeta}_{f}^\top \bbr_j - \hat{\gamma}_{if} }{\hat{\delta}_{if}} + \hat{\alpha}_f  + \hat{\bbeta}_{f}^\top \bbk_j.
\end{equation}
\emph{Known effects} about non-biological variability that we would like to remove are, for instance, the scanner manufacturer (MF) or the magnetic field strength (MFS). 
However, we believe that there are also \emph{unknown effects} that increase variability among sites, which we did not collect explicitly. 
Hence, we propose to adapt an approach that is common in genome-wide association studies (GWAS). 
A major concern in GWAS is population stratification, i.e., the existence of
\revision{unobserved} subpopulations in the sample.
To account for population stratification in the analysis, the addition of principal components (PC) to the regression model has been proposed in EIGENSTRAT~\citep{price2006principal}. 
Principal components are computed on all genetic markers and therefore capture overall variation. 
Following this idea, we compute principal components \emph{across all image features} on the whole dataset to capture generic variation that is not specific to a single brain feature. 
%As we are going to remove variation in the data that is associated with the principal components, 
\revision{Recently, \cite{wang2019blessings} proposed the deconfounder method
that allows inferring multiple causal effects in the presence of unknown
confounders.
Our model is a non-probabilistic version of the deconfounder, which
supports the theoretical connection between our approach and causal inference.
We note that such an approach only helps to identify causal effects} if the neurological process we are studying is believed to only affect a few brain structures.
%\todo[inline]{Mention assumption that this only a good idea if we expect that only a few brain structures are affected by the thing we actually want to study.}
As variables for the remove vector~$\bbr$, we are therefore considering  manufacturer and magnetic field strength, together with the principal component. 

\section{Results}
\label{sec:results}

We first present results for \emph{Name That Dataset} on harmonized features, brain age prediction, and 
a comparison of causal and confounded models with and without harmonization.

\subsection{Name That Dataset with Harmonization}

\begin{figure*}[ht!]
    \centering%
    \includegraphics[width=\textwidth]{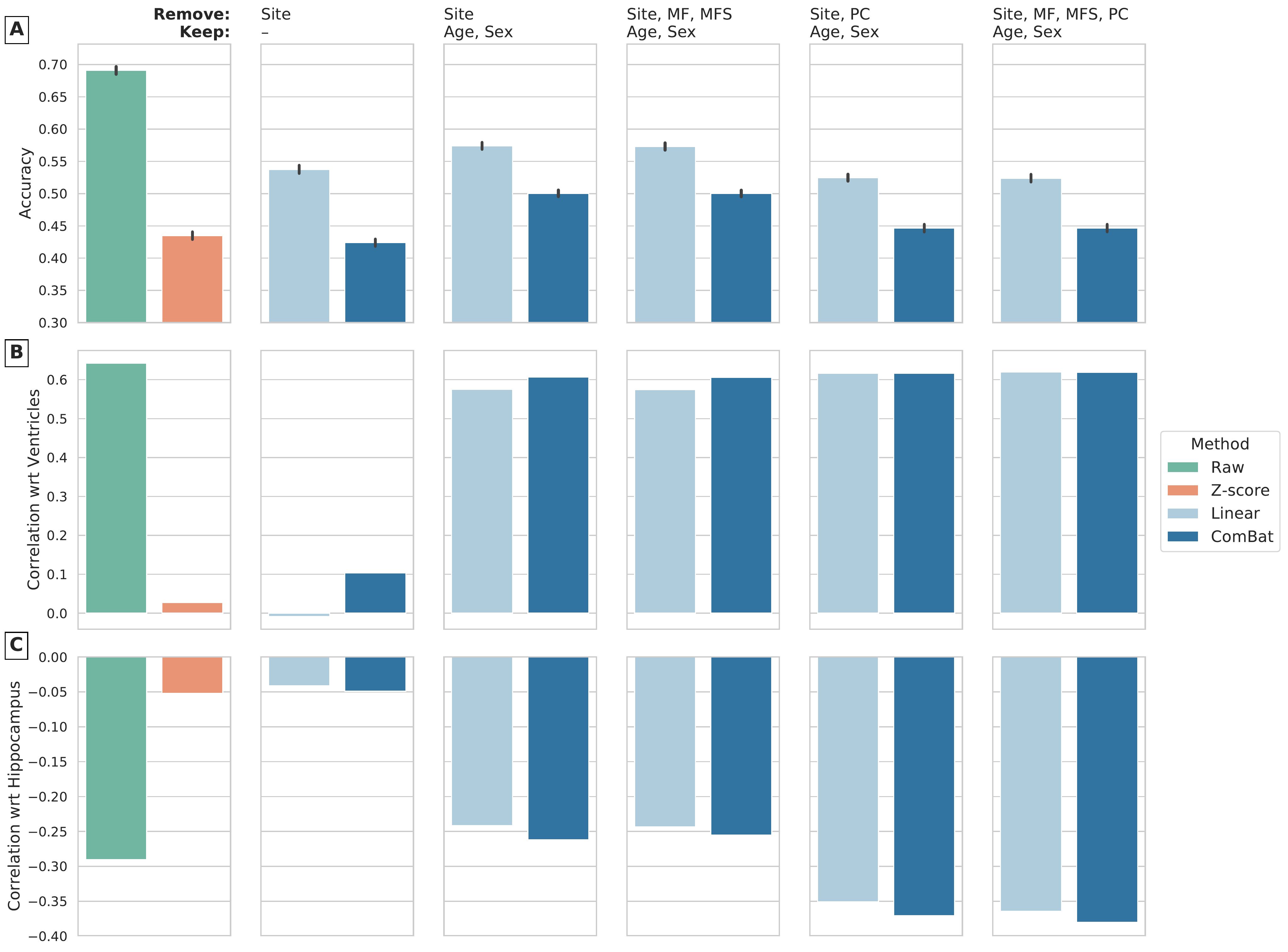}
    \caption{A: Bars show mean dataset classification accuracy with lines indicating standard deviation 
    %\todo[inline]{stddev of what? Add (A), (B), (C) subplots.} 
    for the raw measures and the different harmonization techniques. B: Spearman's rank correlation of lateral ventricles volume with age. C: Spearman's rank correlation of hippocampus volume with age. For the linear and ComBat models, the variables to remove~$\bbr$ and to keep~$\bbk$ are listed above the plots. Variables that we intend on removing are imaging site, manufacturer (MF), magnetic field strength (MFS), principal component (PC).
    \label{fig:siteHarmRes}%
    }%
\end{figure*}

\begin{figure*}
    \centering%
    \includegraphics[scale=0.34]{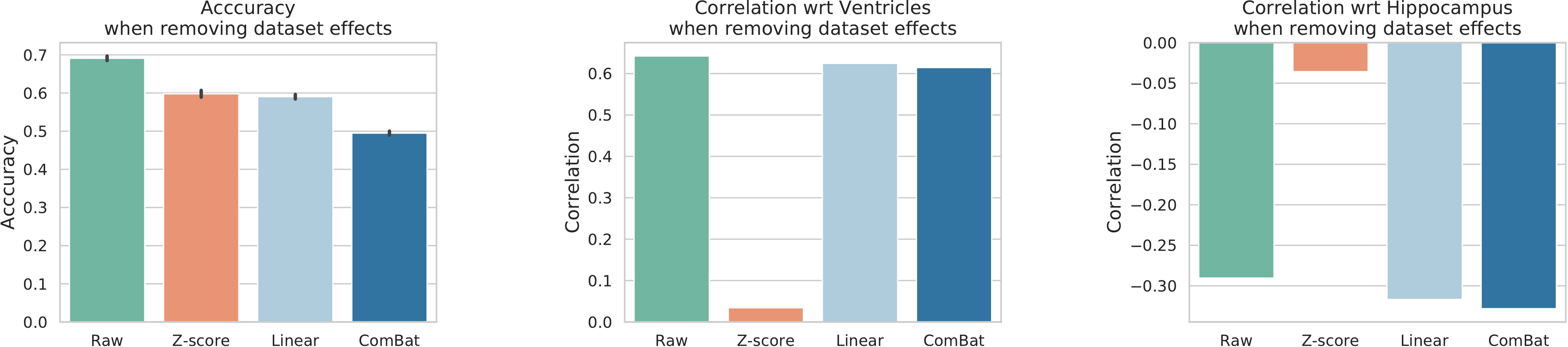}
    \caption{Dataset classification accuracy and Spearman's rank correlation of lateral ventricles and hippocampus with age. In contrast to Fig.~\ref{fig:siteHarmRes}, the dataset is used as grouping variable and not the imaging site. 
    \label{fig:datasetHarmRes}%
    }%
\end{figure*}

Fig.~\ref{fig:siteHarmRes} shows the results of \emph{Name That Dataset}, when the different harmonization techniques were applied. 
We use volume measures, healthy subjects and 70\% training data for these experiments.
Imaging site is used in all linear and ComBat models, further augmented with biological variables age and sex. 
Next, we add non-biological variables that we would like to remove, namely manufacturer (MF), magnetic field strength (MFS), and principal component (PC). 
%We select two PCs, which explain 67\% and 26\% of variance, respectively. 
As noted above, all models that include age have a linear and quadratic aging term. 

While we want to lower the accuracy for predicting the dataset, we also want to keep relevant information. 
As a measure for the latter, we use the fact that ventricles grow with age while
atrophy of the hippocampus increases.
We compute the Spearman's rank correlation of lateral ventricles and hippocampus volumes with age, to account for the non-linear relationship between the volume of brain structures and age.
The objective of this experiment is to validate whether relevant information is retained in the image-derived measurements after harmonization. 
If biologically relevant information is retained,
we should get a positive correlation for lateral ventricles and a negative correlation
for hippocampus volume.
Both volumes are considered relative to the intracranial volume. 
%\todo[inline]{Are your sure? I thought it's absolute volume.}
A trivial solution that would impede predicting the dataset would be to set all variables to constant values, however, this would also remove all relevant information.
Classification accuracies of a random forest are depicted in fig.~\ref{fig:siteHarmRes}A for the original brain measures, Z-Score normalized, and five different variable combinations for the linear and ComBat model.

% \todo[inline]{Add numbers everywhere below.}
Overall, we observe that ComBat is superior to the linear model in this experiment, which indicates the benefit of not only having an additive but also a multiplicative imaging site effect in the model.
Z-Score normalization leads to a steep decrease in classification accuracy from 69\% to 43\%, but also the correlation with age is almost entirely removed (figs.~\ref{fig:siteHarmRes}B and C).
Hence, not only dataset-specific information, but also relevant biological information, is removed.
We observe a similar result for harmonization by residuals based only on imaging site, where ComBat yields a lower classification accuracy than the linear model
(42\% vs.\ 54\%).
Adding age and sex increases the classification accuracy, which
is expected when adding these variables to the residuals computation.
The corresponding ComBat model is closest to the model used by~\cite{fortin2018harmonization} and results in an accuracy of 50\%
and a change in the correlation with age of 0.04.
% For ComBat, the increase is from 42\% to 50\%.
With our extension of including additional variables to remove (MF, MFS, PC), the accuracy decreases to 45\% for ComBat, while at the same time strengthening the correlation with age (0.62 and -0.38).
However, whether the effect of MF and MFS is removed does not change the results
for neither the linear model nor ComBat.
%to remove yields an absolute increase in correlation that surpasses the correlation of the raw measures, as unwanted variation due to scanner differences can obscure the actual relationship. 
%Modeling unknown, global variation with PCs leads to a stronger decrease in classification accuracy than MF and MFS, but also relevant information is reduced, negatively affecting the correlation for hippocampus. 
%The lowest classification accuracies are obtained for the inclusion of all variables, which in case of ComBat, also yields the strongest correlations with age. 
%\todo[inline]{Still higher than Age+Sex baseline in Fig. 2.}
Finally, we note that all classification accuracies are well above 37.9\%, which we reported earlier for the dataset classification task with age and sex, indicating that the features retain dataset-specific information after harmonization. 
%\todo[inline]{45\%, unless the plot is wrong.} indicating that further dataset specific information is reduced. 
%Importantly, the accuracy is decreased while the correlation with age became stronger. 

Fig.~\ref{fig:datasetHarmRes} shows harmonization results for using dataset instead of imaging site as grouping variable. 
Since \emph{Name the Dataset} tries to identify the dataset a scan is part of, it seems natural to harmonize with respect to dataset. 
The figure shows the results for Z-Score normalization, together with the linear and ComBat models that include all variables (age, sex, MF, MFS, PC). %, which performed best in the previous experiment. 
With a classification accuracy of about 50\% for ComBat, this is well above the results for using imaging site. 
Z-Score normalization and the linear model yield accuracies of about 60\%, where the former again removes almost all correlation with age. 
%Compared to about 40\% classification accuracy for using imaging site, the classification accuracy of about 60\% for dataset is much higher. 
%Also, the correlation with age for ventricles and hippocampus stays below the level of the raw data for ComBat. 
%Z-Score normalization again removes almost all correlation. 
We can therefore conclude that operating on the level of imaging site is leading to better results than operating on the dataset level. 
This indicates a high heterogeneity within the included multi-site datasets and the necessity for modeling site-specific variations.

\subsection{Results on Brain Age Prediction}
In a further experiment, we evaluate brain age prediction~\citep{franke2010estimating,cole2017predicting,becker2018gaussian}, i.e., the prediction of a person's age from a brain MRI scan, across datasets. 
The prediction of the brain age is of interest as it was demonstrated that brain age relates to cognitive aging and that the difference to the chronological age is associated to neurodegenerative diseases. 
A prerequisite for age prediction is sufficient data to cover the full age range of interest,
which often requires combining data from multiple datasets. 
However, site-specific characteristics may cause unwanted variation between training and testing sets, as well as heterogeneity in the training set, which can deteriorate prediction results. 
Hence, we examine whether harmonization of the image features can decrease the error. 

\begin{figure}
    \centering%
    \includegraphics[width=0.3\textwidth]{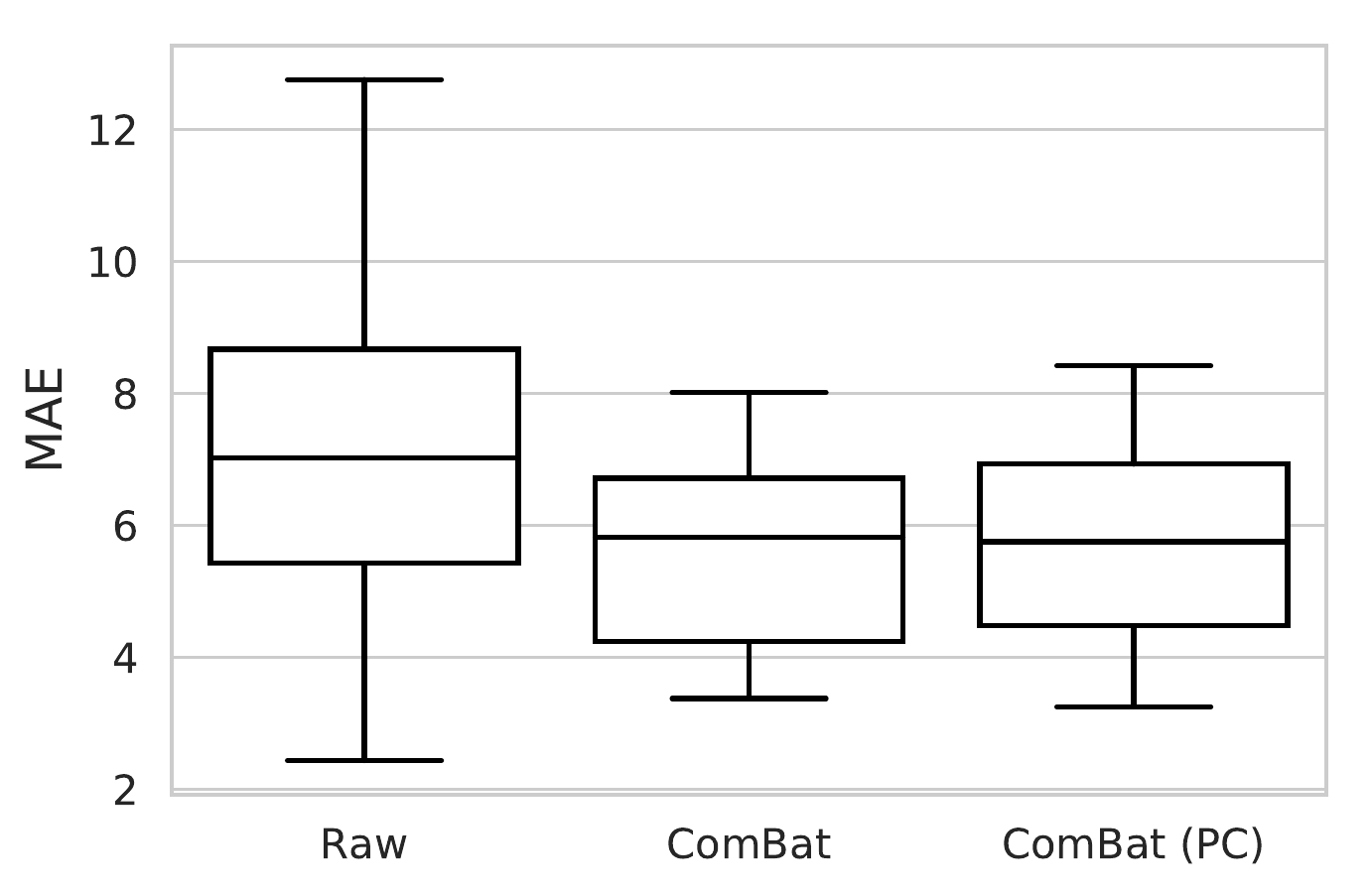}
    \caption{ Boxplot of the mean absolute error (MAE) in age prediction for the raw measurements and the harmonization with ComBat and ComBat with PC. Center lines indicate the median, the boxes extend to the $25^{\text{\tiny th}}$ and $75^{\text{\tiny th}}$ percentiles, and the whiskers reach to the most extreme values. 
    %\todo[inline]{stddev of what? Add (A), (B), (C) subplots.} 
    \label{fig:meaAgePred}%
    }%
\end{figure}

Age prediction is a multivariate regression task, for which we  use random forest regression on brain volumes. 
We only select healthy subjects for this experiment. 
We use a leave-one-dataset-out scheme for the evaluation, i.e., one dataset is selected as test set and the remaining 16 datasets are used for training. 
Fig.~\ref{fig:meaAgePred} shows a boxplot of the mean absolute error in age prediction for the raw measurements, and the harmonized measurements with ComBat and ComBat with PC, respectively. 
\revision{We compute the Wilcoxon signed-rank test between the three approaches yielding $p=0.0064$ between raw and ComBat, $p=0.0197$ between raw and ComBatPC, and $p=0.5171$ between ComBat and ComBatPC; the differences between  the two harmonization techniques are not statistically significant, but the reduction in age prediction errors with respect to raw measures are significant.}
%The reduction in age prediction error between raw image measures and both harmonization approaches is statistically significant with $p < 0.05$; the difference between the two harmonization techniques ($p = 0.47$) is not. %\todo[inline]{Give exact p-values.}

\begin{figure*}[t]
    \centering
    \includegraphics[width=\textwidth]{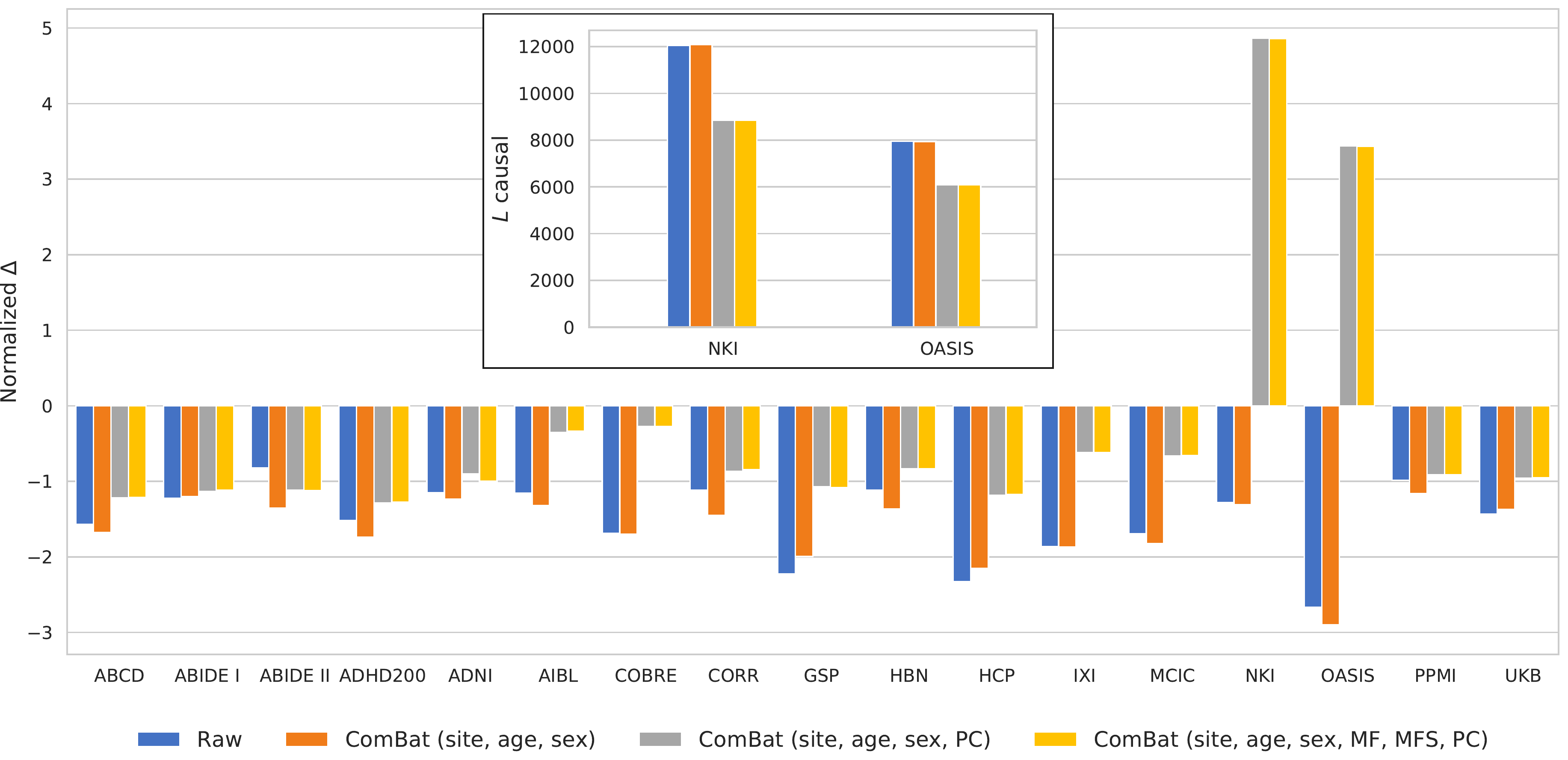}%
    \caption{Results from the causal inference model across all datasets, where the \revision{median} normalized $\Delta$ is plotted. Results are shown for the raw image measurements and the ComBat harmonization with different variable combinations. The \revision{inset} plots $L_\text{ca}$ for NKI and OASIS.
    \revision{Minimum and maximum normalized $\Delta$ values across 10 repetitions are
    available in table~\ref{tab:causal_age_intervals}.}}%
    \label{fig:causal_age}%
\end{figure*}

\begin{table*}[t]
\revision{
\caption{Results from the causal inference model with the variable $Y$ set to ADAS for the raw image measurements and the application of harmonization with different variables. Median values are reported for  $L_\text{ca}$,  $L_\text{co}$, and  $\Delta$, and the minimum and maximum for $\Delta$.}
\begin{center}
  \begin{tabular}{lcccc}
    \toprule
     &  $L_\text{ca}$ &  $L_\text{co}$ & $\Delta$ & $\Delta$ \\
     &   (Median) &  (Median)      &(Median)&  (Range)       \\
    \midrule
    Raw                      & 20,238 & 18,782 & -1,456 & [-1,451; -1,461] \\
    Site                     & 20,221 & 18,789 & -1,432 & [-1,428; -1,435] \\
    Site, Age, Sex           & 20,203 & 18,772 & -1,430 & [-1,429; -1,437] \\
    % Site, Age, Sex, ADAS     & 20159 & 18685 & -1474 & [-1468; -1478] \\
    Site, Age, Sex, PC       & 20,283 & 19,097 & -1,186 & [-1,184; -1,187] \\
    Site, Age, Sex, PC, ADAS & 20,185 & 19,362 &  -823 & [-821; -827] \\
    \bottomrule
    \end{tabular}
\end{center}
\label{tab:causal_adas}
}
\end{table*}%

\subsection{Results on Causality with Harmonization}
We estimate the causal inference model in two experiments. 
In the first one, we select the ADNI dataset and set $Y$ to the Alzheimer's Disease Assessment Scale Cognitive Subscale 13 (ADAS) score. %\todo[inline]{Citation needed?}
ADAS is one of the most widely used cognitive scales and assesses the severity of cognitive symptoms of dementia~\citep{kueper2018alzheimer}. 
Instead of directly working with diagnosis, ADAS is a continuous variable and can therefore directly be used in our causal inference framework.  
For the variable $\mathbf{X}$, we use age, age squared, sex, and the following brain volumes that have previously been associated with dementia: hippocampus, amygdala, lateral ventricles, inferior lateral ventricles, and third ventricle~\citep{apostolova2012hippocampal}.
The model complexities for causal and confounded models, together with the difference $\Delta$ for the raw volume measurements is reported in Table~\ref{tab:causal_adas}.
Next, we evaluate different harmonization methods for the volume measures, where we focus on ComBat. 
\revision{When only using site, or site in combination with age and sex,
the intervals for $\Delta$ overlap, which is evidence for both working equally well.
The inclusion of PC yields a higher median $\Delta$ and its interval is cleary
seperated from the results above, which indicates a less confounded model.
Adding ADAS in the harmonization to the variables to keep~($\bbk$),
further reduces confounding bias significantly,
as is evident from non-overlapping intervals for $\Delta$.}

%\todo[inline]{Add some sort of explanation why we need to use age here.}
In the second experiment, we want to include all datasets in the study. 
To this end, we set $Y$ to age and $\mathbf{X}$ to the following brain volumes: hippocampus, lateral ventricles, amygdala, inferior lateral ventricles,  third ventricle, putamen, pallidum, caudate, thalamus proper, cerebellum cortex, cerebellum white matter, and brain stem, together with sex. 
The results per dataset are shown in Fig.~\ref{fig:causal_age}, where we present the normalized $\Delta$ for better comparison across datasets. 
\revision{Next to the median values displayed in Fig.~\ref{fig:causal_age}, we list the range in table~\ref{tab:causal_age_intervals}.}
We report results for raw image measures and the harmonization with ComBat. 
As variables, we use (i) site, age, sex, (ii) site, age, sex, PC, and (iii) site, age, sex, MF, MFS, PC.
For (i), which corresponds to the original \revision{version} of ComBat, we observe similar values to the raw measures. 
After including PC, we observe a substantial increase in $\Delta$ for almost all datasets. 
Notably, for NKI and OASIS, the difference gets positive, which indicates a causal relationship. 
To investigate this further, we plot $L_{\text{ca}}$ for NKI and OASIS in the \revision{inset} in Fig.~\ref{fig:causal_age}, which shows the large drop in complexity for the causal model after harmonization with PC.  
With the complexity of the confounded model changing only little, it is the change of the causal model that yields the positive $\Delta$ and therefore evidence for a causal relationship.

\section{Discussion}

\subsection{\emph{Name That Dataset}}
In one of the largest studies to date, we have examined bias across 17 neuroimaging datasets. 
Our experiments for \emph{Name That Dataset} demonstrated that the dataset can be predicted with more than 70\% accuracy. %\todo[inline]{On healthy subjects?}
In comparison, a classifier that operates on age and sex stays below 38\%. 
This is evidence that image features contain, next to subject-specific information, also rich dataset-specific information. % (also referred to as biological versus non-biological variation). 
While we are not aware of previous work on predicting the dataset from neuroimaging data, prior studies have reported the high intra-subject variability across scanners, no matter whether FreeSurfer~\citep{jovicich2009mri}, FSL~\citep{nugent2013automated,suckling2012neuro}, or BrainVisa~\citep{shokouhi2011assessment} were used for segmentation, which is also pointing to scanner-specific information in the data. 
%In the literature, multiple studies  demonstrated the high inter-scanner variability, no matter whether FreeSurfer~\citep{jovicich2009mri}, FSL~\citep{nugent2013automated,suckling2012neuro}, or BrainVisa~\citep{shokouhi2011assessment} were used for segmentation. 

Instead of predicting the dataset, a similar experiment would be to predict  manufacturer or magnetic field strength of a scan. 
We believe that dataset prediction offers two advantages. 
First, there is a severe class imbalance with almost 80\% of scans acquired on a Siemens scanner and 84\% on 3.0T, yielding already high classification accuracies with na{\"i}ve approaches.
Second, there are more non-biological effects than just manufacturer and field strength that impact a scan; by using the dataset as label, we can lump them all together, without explicitly knowing them. 
However, we also need to account for biological variation between datasets, which could be considered as a lower bound for the classification accuracy, as we would like to retain this information. 
In our experiments, we used \revision{a classification model based} on the variables age and sex for this purpose. 
Another alternative to \emph{Name That Dataset} would be to \emph{Name That Imaging Site}. 
With 210 sites in our data, this would substantially increase the complexity of the classification task and complicate the interpretation of results with the confusion matrix. 
In addition, there is a wide variation in the size of sites, ranging from a few subjects per site to more than 1,000 for HCP. 
Hence, \emph{Name That Dataset} is the most reasonable choice in our experiments, but operating on sites could be an attractive alternative in other situations.

\subsection{Dataset Harmonization}
We have investigated several techniques for dataset harmonization and evaluated \revision{whether} they can reduce the accuracy of predicting the dataset or whether they can yield a causal relationship. 
Harmonization is a trade-off between removing unwanted variation, likely associated to scanner, and keeping relevant information about the research question. 
Hence, we need to be cautious when harmonizing image features. 
As shown in our results for \emph{Name That Dataset}, Z-score normalization and the harmonization with imaging site can only drastically decrease the classification accuracy. 
However, the low correlation of age with hippocampus and ventricles volumes after harmonization illustrates
that all the biologically relevant information has been removed too. 
Adding more supervision to the process is therefore necessary, which we achieved by adding variables whose association we want to keep in the data (e.g.\ age and sex). 
Results from our correlation analyses with age show the effectiveness of such an approach, but also that the dataset prediction accuracy increases compared to only using site.

As a novel contribution to harmonization, we proposed to add variables to the model, whose influence we would like to remove. 
In particular, we proposed to compute principal components across all image features, and add them as variables to remove. 
This approach is inspired from GWAS, where principal components across genetic markers are added to a model of association to account for stratification of the population. 
In contrast, we use principal components in a regression model to harmonize imaging data by computing the residuals. 
Next to PC, our technique can also be used to explicitly remove the effect of measured variables;
we studied scanner manufacturer and magnetic field strength. 
In combination with imaging site as grouping variable, their removal had only small impact. 
This is most likely due to manufacturer and field strength being constant within an imaging site, so that the information provided by these variables is already accounted for by imaging site. 
Manufacturer and field strength have a stronger influence on the results when dataset is used as grouping variable. 
%Most likely, this is due to using imaging site as grouping variable: since each site has the same manufacturer and magnetic field strength this is already accounted for. 
%\todo[inline]{I don't understand this point.}

The prediction of the brain age is a typical example, where the combination of multiple dataset is required to cover a wide age range in the training dataset. 
The results of our leave-one-dataset-out experiment demonstrated that harmonization yields a significant reduction in the age prediction error. 
As age prediction has been proposed as a diagnostic biomarker for many diseases, like AD, HIV, and schizophrenia~\citep{cole2017predicting}, accounting for dataset bias may further increase its sensitivity.

For the computation of residuals in the harmonization, we worked with linear regression models and ComBat, where ComBat has an additional site-specific scaling factor and uses empirical Bayes for inference. 
Throughout all our experiments, we noticed advantages of ComBat in comparison to the linear model. 
This confirms previous results from~\cite{fortin2018harmonization}  for harmonizing cortical thickness measures with ComBat  also for the harmonization of volume measurements. 
We have focused on harmonization approaches that operate on image-derived features, while previous approaches harmonized image intensity values to make them comparable across studies, including histogram matching~\citep{nyul2000new}, WhiteStripe~\citep{shinohara2014statistical}, and RAVEL~\citep{fortin2016removing}.
\revision{As an alternative to accounting for population stratification with PCs added to the model, mixed linear models have been proposed for genetic association studies~\citep{yang2014mixed}, where the random effect accounts for stratification. 
We could adopt a similar approach to model unknown effects in image features. 
However, since random effects are not explicitly estimated, correcting the image features in a harmonization step is not directly possible, but rather an integrated model would be required.}
%Such an approach could also be adopted for harmonization of imaging features, but it does not allow for updating the features (regress out), but rather requires the analysis in the linear mixed effects model. The application of more complex models, like random forests, would not be directly possible.}

\subsection{Causal Inference for Identifying Confounding}
%\todo[inline]{I would move this section to the end of results section to have the same order as our experiments.}
Confounding bias can lead to spurious correlations and therefore wrong conclusions about cause and effect. 
The assumption of causal sufficiency, i.e., knowing all confounding variables, is often violated in practice. 
Hence, we presented a causal inference framework that considers \emph{unknown confounders}. 
%It is based on determining whether a confounded or causal relationship is more likely. 
In contrast, prior work  in neuroimaging~\citep{dukart2011age,linn2016addressing,rao2017predictive} mainly considered the known confounders age and sex. 
We believe that our approach can present an interesting new analysis for neuroimaging data, e.g.,  for evaluating methods for harmonization  or  bias reduction. 

%Prior studies on confounding bias in neuroimaging~\citep{dukart2011age,linn2016addressing,rao2017predictive} have mainly focused on confounding within a dataset, with age and sex as potential confounders, but not across datasets. 

Our results indicate that most relationships that we investigated were confounded rather than causal. 
This is unsurprising, given the inherent complexity of neuroimaging studies related to scanning, image analysis, and study design. 
On the positive side, techniques for harmonization can reduce confounding and, in some cases, even yield causal relationships. 
For both causal experiments, we \revision{observed that accounting for unknown confounders in ComBat via principal components is preferred.
In addition, the experiment on the ADNI dataset revelead that
including the target variable (ADAS) as variable to keep,
results in an improved harmonization that reduces confounding significantly.
However, the relationship remains non-causal ($\Delta <0$). We believe that morphology alone may be insufficient, and that additional, multi-modal measures would be necessary to establish a causal model for ADAS.}

As mentioned earlier, inferring causality from observational data is challenging and only feasible under certain assumptions as stated in section~\ref{sec:causal}.
\revision{These assumptions heavily rely on domain knowledge
and may not hold for other applications.
For instance, we assumed a linear relationship between continuous-valued variables,
which suggested the use of linear regression as causal model and probabilistic PCA
as confounded model. If non-linearity is required, more advanced
models such as Gaussian Process regression and
probabilistic non-linear PCA
could be used~\citep{rasmussen2006gaussian,Lawrence2005}.
Similarly, if data comprises discrete variables, appropriate models need to be selected,
such as Poisson regression as casual model and Gamma-Poisson factorization
as confounded model~\citep{Canny2004}.
All of these models would fit the setting depicted in Fig.~\ref{fig:causalDiag}
and allow estimation of $\Delta$ as described in section~\ref{sec:causal}.
%so that the results have to be interpreted with caution.
In addition to relying on untestable assumptions, causal inference
from observational data is challenging, because we cannot know what the true
causal effects are nor their effect size, which makes a quantitative evaluation impossible.}
As such, results need to be interpreted with caution and in light of the incorporated prior knowledge. 
Yet, the ability to assess  confounding in the model can be very helpful, not just for the development of harmonization tools, but more generally for inferring knowledge from neuroimaging studies. 
%determining whether the relationship between variables is 

\section{Conclusion}
Bias is a complex and challenging topic in neuroimaging that will become more prevalent in the future with the translation of results in the clinic and the surge of mega-analyses. 
%the combination of large datasets and the translation to the clinic. 
%We defined various forms of bias common to neuroimaging data.
Based on data with more than 35,000 individuals, we have demonstrated that simply pooling scans from distinct studies can introduce substantial bias that would be passed on to a machine learning model trained on the pooled data.
First, we showed that it is possible to correctly identify the dataset that a scan is part of with more than 70\% accuracy. 
Second, we introduced a novel approach for differentiating causal from confounded relationships based on causal inference. 
Importantly, the confounder was modeled as unknown variable, which is helpful in complex neuroimaging studies, where the assumption of causal sufficiency is challenging to fulfill. % all influences are not known nor measured. 
Third, we studied multi-site harmonization techniques and evaluated their effectiveness of reducing bias. 
We have proposed a harmonization method that extends ComBat for the inclusion of additional variables to remove, where the integration of principal components, to capture generic variation, led to the best results. 

%We estimated the strength of the neurobiological causal model -- age and sex influence
%a brain structure's volume -- versus the confounded model -- age, sex,
%and volume are influenced by latent confounders -- for 15 datasets and 22 brain structures.
%Results yielded large differences in the causal strength across datasets and brain structures.
%These results are specific to our assumptions for causal and confounding model, where other choices are possible and may yield different conclusions.

Overall, we believe that the growing amount of medical images necessitates novel methods for handling bias in datasets and image-derived features. 
As bias is in its core a causal concept, methods from the growing field of causal inference may be particularly promising to yield new insights. 
%, for which we have proposed a causal model in this work.
%Finally, we note that our approach is not restricted to a
%biology-derived causal model, but could also be used to estimate the
%causal effect for other relationships, such as the effect of
%magnetic field strength on signal-to-noise ratio.

\section*{Acknowledgements}
This research was supported by the Bavarian State Ministry of Science and the Arts and coordinated by the Bavarian Research Institute for Digital Transformation (bidt).
The authors gratefully acknowledge the Leibniz Supercomputing Centre for providing computing time on its Linux-Cluster.

We acknowledge the following initiatives for providing data. 

ABCD: Data used in the preparation of this article were obtained from the Adolescent Brain
Cognitive Development (ABCD) Study (https://abcdstudy.org), held in the NIMH Data Archive
(NDA). This is a multisite, longitudinal study designed to recruit more than 10,000 children aged
9-10 years and follow them over 10 years into early adulthood. The ABCD Study is supported
by the National Institutes of Health and additional federal partners under award numbers
U01DA041022, U01DA041028, U01DA041048, U01DA041089, U01DA041106, U01DA041117,
U01DA041120, U01DA041134, U01DA041148, U01DA041156, U01DA041174, U24DA041025,
U01DA041093, U24DA041123, and U24DA041147. A full list of supporters is available at
https://abcdstudy.org/nih-collaborators. A listing of participating sites and a complete listing of
the study investigators can be found at https://abcdstudy.org/principal-investigators.html. ABCD
consortium investigators designed and implemented the study and/or provided data but did not
necessarily participate in analysis or writing of this report. This manuscript reflects the views of
the authors and may not reflect the opinions or views of the NIH or ABCD consortium
investigators. 

ADNI: Data collection and sharing for this project was funded by the Alzheimer's Disease
Neuroimaging Initiative (ADNI) (National Institutes of Health Grant U01 AG024904) and
DOD ADNI (Department of Defense award number W81XWH-12-2-0012). ADNI is funded
by the National Institute on Aging, the National Institute of Biomedical Imaging and
Bioengineering, and through generous contributions from the following: Alzheimer's
Association; Alzheimer's Drug Discovery Foundation; Araclon Biotech; BioClinica, Inc.;
Biogen Idec Inc.; Bristol-Myers Squibb Company; Eisai Inc.; Elan Pharmaceuticals, Inc.; Eli
Lilly and Company; EuroImmun; F. Hoffmann-La Roche Ltd and its affiliated company
Genentech, Inc.; Fujirebio; GE Healthcare; ; IXICO Ltd.; Janssen Alzheimer Immunotherapy
Research \& Development, LLC.; Johnson \& Johnson Pharmaceutical Research \&
Development LLC.; Medpace, Inc.; Merck \& Co., Inc.; Meso Scale Diagnostics,
LLC.; NeuroRx Research; Neurotrack Technologies; Novartis Pharmaceuticals
Corporation; Pfizer Inc.; Piramal Imaging; Servier; Synarc Inc.; and Takeda Pharmaceutical
Company. The Canadian Institutes of Health Research is providing funds to support ADNI
clinical sites in Canada. Private sector contributions are facilitated by the Foundation for the
National Institutes of Health (www.fnih.org). The grantee organization is the Northern
California Institute for Research and Education, and the study is coordinated by the
Alzheimer's Disease Cooperative Study at the University of California, San Diego. ADNI
data are disseminated by the Laboratory for Neuro Imaging at the University of Southern
California.

COBRE: Data was downloaded from the COllaborative Informatics and Neuroimaging Suite Data Exchange tool (COINS; http://coins.mrn.org/dx) and data collection was performed at the Mind Research Network, and funded by a Center of Biomedical Research Excellence (COBRE) grant 5P20RR021938/P20GM103472 from the NIH to Dr. Vince Calhoun.

MIND: Data used in the preparation of this work were obtained from the Mind Clinical Imaging Consortium database through the Mind Research Network (www.mrn.org). The MCIC project was supported by the Department of Energy under Award Number DE-FG02-08ER64581. MCIC is the result of efforts of co-investigators from University of Iowa, University of Minnesota, University of New Mexico, Massachusetts General Hospital.

GSP: Data were provided in part by the Brain Genomics Superstruct Project of Harvard University and the Massachusetts General Hospital, (Principal Investigators: Randy Buckner, Joshua Roffman, and Jordan Smoller), with support from the Center for Brain Science Neuroinformatics Research Group, the Athinoula A. Martinos Center for Biomedical Imaging, and the Center for Human Genetic Research. 20 individual investigators at Harvard and MGH generously contributed data to the overall project.

HCP:  Data were provided in part by the Human Connectome Project, WU-Minn Consortium (Principal Investigators: David Van Essen and Kamil Ugurbil; 1U54MH091657) funded by the 16 NIH Institutes and Centers that support the NIH Blueprint for Neuroscience Research; and by the McDonnell Center for Systems Neuroscience at Washington University

OASIS: Data were provided in part by OASIS Principal Investigators: D. Marcus, R, Buckner, J, Csernansky J. Morris; P50 AG05681, P01 AG03991, P01 AG026276, R01 AG021910, P20 MH071616, U24 RR021382.

PPMI: Data used in the preparation of this article were obtained from the Parkinson's Progression Markers Initiative (PPMI) database (www.ppmi-info.org/data). For up-to-date information on the study, visit www.ppmi-info.org.  PPMI a public-private partnership is funded by the Michael J. Fox Foundation for Parkinson's Research and funding partners, including [list the full names of all of the PPMI funding partners found at www.ppmi-info.org/fundingpartners].

UKB: This research has been conducted using the UK Biobank Resource under the application number 34479. 

\section{References}

\bibliographystyle{elsarticle-harv}
\bibliography{jab_bib}

\appendix
\section{Additional Results}
\begin{table*}\centering
\revision{
  \caption{\label{tab:causal_age_intervals}%
  Median, minimum, and maximum normalized $\Delta$ across 10 repetitions
  from the causal inference model across all datasets (cf.~Fig.~\ref{fig:causal_age}).%
  }
  \small
  \begin{tabular}{lcccc}
    \toprule
    Dataset & Raw & ComBat           & ComBat               & ComBat \\
    {}      &     & (site, age, sex) & (site, age, sex, PC) & (site, age, sex, MF, MFS, PC) \\
    \midrule
    ABCD     &  -1.571  [-1.572; -1.570] &  -1.676  [-1.677; -1.675] &    -1.219  [-1.220; -1.218] &             -1.214  [-1.216; -1.213] \\
    ABIDE I  &  -1.226  [-1.229; -1.224] &  -1.202  [-1.205; -1.201] &    -1.135  [-1.139; -1.133] &             -1.119  [-1.121; -1.117] \\
    ABIDE II &  -0.824  [-0.825; -0.823] &  -1.354  [-1.356; -1.352] &    -1.119  [-1.125; -1.118] &             -1.120  [-1.121; -1.117] \\
    ADHD200  &  -1.520  [-1.527; -1.519] &  -1.740  [-1.742; -1.738] &    -1.285  [-1.293; -1.282] &             -1.274  [-1.276; -1.273] \\
    ADNI     &  -1.153  [-1.155; -1.152] &  -1.235  [-1.238; -1.233] &    -0.901  [-0.902; -0.898] &             -1.000  [-1.003; -0.998] \\
    AIBL     &  -1.154  [-1.154; -1.154] &  -1.318  [-1.318; -1.318] &    -0.354  [-0.354; -0.353] &             -0.336  [-0.337; -0.336] \\
    COBRE    &  -1.691  [-1.691; -1.690] &  -1.701  [-1.701; -1.700] &    -0.276  [-0.277; -0.276] &             -0.275  [-0.275; -0.274] \\
    CORR     &  -1.117  [-1.121; -1.116] &  -1.449  [-1.451; -1.446] &    -0.870  [-0.873; -0.868] &             -0.847  [-0.851; -0.845] \\
    GSP      &  -2.228  [-2.232; -2.225] &  -1.995  [-1.998; -1.993] &    -1.073  [-1.079; -1.071] &             -1.081  [-1.085; -1.079] \\
    HBN      &  -1.115  [-1.115; -1.114] &  -1.363  [-1.365; -1.362] &    -0.834  [-0.836; -0.832] &             -0.833  [-0.838; -0.832] \\
    HCP      &  -2.330  [-2.337; -2.327] &  -2.150  [-2.156; -2.148] &    -1.184  [-1.186; -1.182] &             -1.172  [-1.174; -1.170] \\
    IXI      &  -1.866  [-1.867; -1.865] &  -1.866  [-1.868; -1.863] &    -0.620  [-0.621; -0.619] &             -0.619  [-0.621; -0.618] \\
    MCIC     &  -1.694  [-1.694; -1.694] &  -1.821  [-1.821; -1.821] &    -0.662  [-0.662; -0.662] &             -0.661  [-0.661; -0.660] \\
    NKI      &  -1.280  [-1.281; -1.279] &  -1.307  [-1.308; -1.307] &       4.864  [4.862; 4.866] &                4.863  [4.861; 4.864] \\
    OASIS    &  -2.664  [-2.664; -2.664] &  -2.901  [-2.901; -2.901] &       3.439  [3.439; 3.439] &                3.433  [3.433; 3.433] \\
    PPMI     &  -0.986  [-0.987; -0.986] &  -1.161  [-1.162; -1.161] &    -0.913  [-0.913; -0.912] &             -0.915  [-0.915; -0.914] \\
    UKB      &  -1.432  [-1.433; -1.432] &  -1.374  [-1.374; -1.372] &    -0.955  [-0.956; -0.955] &             -0.952  [-0.953; -0.951] \\
    \bottomrule
    \end{tabular}
    }
\end{table*}

\end{document}